\documentclass[twoside]{article}
\usepackage{aistats2014}

\usepackage{algorithm}
\usepackage{algorithmic}
\usepackage{natbib}
\usepackage{amsmath}
\usepackage{amssymb}
\usepackage{latexsym}
\usepackage{microtype}
\usepackage{epsfig}
\usepackage{fixmath}
\usepackage{titlesec}
\setcitestyle{round}
\usepackage{framed}

\newcommand{\cL}{ {\cal L}}
\newcommand{\E}{\textrm{E}}
\newcommand{\myeqp}[1]{Eq.~\ref{eq:#1}}

\newcommand{\myfig}[1]{Figure~\ref{fig:#1}}
\newcommand{\g}{\,|\,}

\begin{document}

\runningtitle{Black Box Variational Inference}

\twocolumn[

\aistatstitle{Black Box Variational Inference}

\aistatsauthor{ Rajesh Ranganath  \And Sean Gerrish \And David M. Blei}
\aistatsaddress{ Princeton University, 35 Olden St., Princeton, NJ 08540 \\
{\tt $\{$rajeshr,sgerrish,blei$\}$ AT cs.princeton.edu}}
]

\begin{abstract}
  Variational inference has become a widely used method to approximate
  posteriors in complex latent variables models.  However, deriving a
  variational inference algorithm generally requires significant
  model-specific analysis, and these efforts can hinder and deter us
  from quickly developing and exploring a variety of models for a
  problem at hand.  In this paper, we present a ``black box''
  variational inference algorithm, one that can be quickly applied to
  many models with little additional derivation.  Our method is based
  on a stochastic optimization of the variational objective where the
  noisy gradient is computed from Monte Carlo samples from the
  variational distribution.  We develop a number of methods to reduce
  the variance of the gradient, always maintaining the criterion that
  we want to avoid difficult model-based derivations.  We evaluate our
  method against the corresponding black box sampling based
  methods. We
  find that our method reaches better predictive likelihoods much faster
  than sampling methods.
  Finally, we demonstrate that Black Box Variational
  Inference lets us easily explore a wide space of models by quickly
  constructing and evaluating several models of longitudinal
  healthcare data.
\end{abstract}

\section{Introduction}

Probabilistic models with latent variables have become a mainstay in
modern machine learning applications. With latent variables models, we
posit a rich latent structure that governs our observations, infer
that structure from large data sets, and use our inferences to
summarize observations, draw conclusions about current data, and make
predictions about new data.  Central to working with latent variable
models is the problem of computing the posterior distribution of the
latent structure. For many interesting models, computing the
posterior exactly is intractable: practitioners must resort to
approximate methods.

One of the most widely used methods for approximate posterior
estimation is variational inference~\citep{Wainwright:2008,
  Jordan:1999}.  Variational inference tries to find the member of a
family of simple probability distributions that is closest (in KL
divergence) to the true posterior distribution.

For a specific class of models, those where the conditional
distributions have a convenient form (and where a convenient
variational family exists), this optimization can be carried out with
a closed-form coordinate ascent algorithm~\citep{Ghahramani:2001}.
For generic models and arbitrary variational families, however, there
is no closed-form solution: computing the required expectations
becomes intractable.  In these settings, practitioners have resorted
to model-specific algorithms \citep{Jaakkola:1996, Blei:2007,
  Braun:2010} or generic algorithms that require model specific
computations \citep{Knowles:2011, Wang:2012, Paisley:2012}.

Deriving these algorithms on a model-by-model basis is tedious work.
This hinders us from rapidly exploring modeling assumptions when
solving applied problems, and it makes variational methods on
complicated distributions impractical for many practitioners.  Our
goal in this paper is to develop a ``black box'' variational inference
algorithm, a method that can be quickly applied to almost any model
and with little effort.  Our method allows practitioners to quickly
design, apply, and revise models of their data, without painstaking
derivations each time they want to adjust the model.

Variational inference methods frame a posterior estimation problem as
an optimization problem, where the parameters to be optimized adjust a
variational ``proxy'' distribution to be similar to the true
posterior.  Our method rewrites the gradient of that objective as the
expectation of an easy-to-implement function $f$ of the latent and
observed variables, where the expectation is taken with respect to the
variational distribution; and we optimize that objective by sampling
from the variational distribution, evaluating the function $f$, and
forming the corresponding Monte Carlo estimate of the gradient.  We
then use these stochastic gradients in a stochastic optimization
algorithm to optimize the variational parameters.

From the practitioner's perspective, this method requires only that
he or she write functions to evaluate the model
log-likelihood.  The remaining calculations (sampling from the
variational distribution and evaluating the Monte Carlo estimate) are
easily put into a library to be shared across models, which means our
method can be quickly applied to new modeling settings.

We will show that reducing the variance of the gradient estimator is
essential to the fast convergence of our algorithm.  We develop
several strategies for controlling the variance.  The first is based on
Rao-Blackwellization \citep{Casella:1996}, which exploits the
factorization of the variational distribution.  The second is based on
control variates ~\citep{Ross:2002, Paisley:2012}, using the log probability of the
variational distribution.  We emphasize that these variance reduction
methods preserve our goal of black box inference because they do not
require computations specific to the model.

Finally, we show how to use recent innovations in variational
inference and stochastic optimization to scale up and speed up our
algorithm.  First, we use adaptive learning rates~\citep{Duchi:2011}
to set the step size in the stochastic optimization.  Second, we
develop generic stochastic variational inference~\citep{Hoffman:2013},
where we additionally subsample from the data to more cheaply compute
noisy gradients.  This innovates on the algorithm
of~\citet{Hoffman:2013}, which requires closed form coordinate updates
to compute noisy natural gradients.

We demonstrate our method in two ways.  First, we compare our method
against Metropolis-Hastings-in-Gibbs \citep{Bishop:2006}, a sampling
based technique that requires similar effort on the part of the
practitioner.  We find our method reaches better predictive
likelihoods much faster than sampling methods.  Second, we use our
method to quickly build and evaluate several models of longitudinal
patient data.  This demonstrates the ease with which we can now
consider models generally outside the realm of variational methods.

\paragraph{Related work.}  There have been several lines of work that
use sampling methods to approximate gradients in variational
inference. \citet{Wingate:2013} have independently considered a
similar procedure to ours, where the gradient is construed as an
expectation and the KL is optimized with stochastic optimization.
They too include a term to reduce the variance, but do not describe
how to set it.  We further innovate on their approach with Rao-Blackwellization, 
specified control variates, adaptive learning rates,
and data subsampling.  \citet{Salimans:2012} provide a framework based
on stochastic linear regression. Unlike our approach, their method
does not generalize to arbitrary approximating families and requires
the inversion of a large matrix that becomes impractical in high
dimensional settings.  \citet{Kingma:2013} provide an alternative
method for variational inference
through a reparameterization of the variational distributions. 
In constrast to our approach, their algorithm is limited to
only continuous latent variables. \citet{Carbonetto:2009}
present a stochastic optimization scheme for moment estimation
based on the specific form of the variational objective when
both the model and the approximating family are in the same 
exponential family. This differs from our more general modeling
setting where latent variables may be outside of the exponential family. 
Finally, \citet{Paisley:2012} use Monte Carlo
gradients for difficult terms in the variational objective and also
use control variates to reduce variance.  However, theirs is not a
black-box method. Both the objective function and control variates
they propose require model-specific derivations.

\section{Black Box Variational Inference}

Variational inference transforms the problem of approximating a
conditional distribution into an optimization
problem~\citep{Jordan:1999, Bishop:2006, Wainwright:2008}.  The idea is to posit a
simple family of distributions over the latent variables and find the
member of the family that is closest in KL divergence to the conditional
distribution.

In a probabilistic model, let $x$ be observations, $z$ be latent
variables, and $\lambda$ the free parameters of a variational
distribution $q(z \g \lambda)$.  Our goal is to approximate $p(z \g
x)$ with a setting of $\lambda$.  In variational inference we optimize
the Evidence Lower BOund (ELBO),
\begin{align} \label{eq:objective}
  \cL(\lambda) \triangleq \E_{q_\lambda(z)}[\log p(x,z) - \log q(z)].
\end{align}
Maximizing the ELBO is equivalent to minimizing the KL
divergence~\citep{Jordan:1999, Bishop:2006}.  Intuitively, the first
term rewards variational distributions that place high mass on
configurations of the latent variables that also explain the
observations; the second term rewards variational distributions that
are entropic, i.e., that maximize uncertainty by spreading their mass
on many configurations.

Practitioners derive variational algorithms to maximize the ELBO
over the variational parameters by expanding the expectation in
\myeqp{objective} and then computing gradients to use in an
optimization procedure.  Closed form coordinate-ascent updates are
available for conditionally conjugate exponential family models~
\citep{Ghahramani:2001}, where the distribution of each latent
variable given its Markov blanket falls in the same family as the
prior, for a small set of variational families.  However, these
updates require analytic computation of various expectations for each
new model, a problem which is exacerbated when the variational family
falls outside this small set.  This leads to tedious bookkeeping and
overhead for developing new models.

We will instead use stochastic optimization to maximize the ELBO.  In
stochastic optimization, we maximize a function using noisy estimates
of its gradient~\citep{Robbins:1951,Kushner:1997,Bottou:2004a}.  We
will form the derivative of the objective as an expectation with
respect to the variational approximation and then sample from the
variational approximation to get noisy but unbiased gradients, which
we use to update our parameters.
For each sample, our noisy gradient requires evaluating the joint
distribution of the observed and sampled variables, the variational
distribution, and the gradient of the log of the variational
distribution.  This is a black box method in that the gradient of the
log of the variational distribution and sampling method can be derived
once for each type of variational distribution and reused for
many models and applications.

\paragraph{Stochastic optimization.}  We first review stochastic
optimization. Let $f(x)$ be a function to be maximized and $h_t(x)$ be
the realization of a random variable $H(x)$ whose expectation is the
gradient of $f(x)$.  Finally, let $\rho_t$ be the learning rate.
Stochastic optimization updates $x$ at the $t$th iteration with
\begin{equation*}
  x_{t+1} \gets x_{t} + \rho_t h_t(x_t).
\end{equation*}
This converges to a maximum of $f(x)$ when the learning rate schedule
follows the Robbins-Monro conditions,
\begin{eqnarray*}
  \textstyle \sum_{t=1}^\infty \rho_t &=& \infty \\
  \textstyle \sum_{t=1}^\infty \rho_t^2 &<& \infty. \label{eq:rm}
\end{eqnarray*}
Because of its simplicity, stochastic optimization is widely used in
statistics and machine learning.

\paragraph{A noisy gradient of the ELBO.}  To optimize the ELBO with
stochastic optimization, we need to develop an unbiased estimator of
its gradient which can be computed from samples from the variational
posterior. To do this, we write the gradient of the ELBO
(\myeqp{objective}) as an expectation with respect to the variational
distribution,
\begin{align}
\nabla_\lambda \cL = \E_q [\nabla_\lambda \log q(z | \lambda) (\log p(x, z) - \log q(z | \lambda))].
\label{eq:gradient}
\end{align}
The derivation of
\myeqp{gradient} can be found in the appendix.  Note that in
statistics the gradient $\nabla_\lambda \log q(z | \lambda)$ of the
log of a probability distribution is called the score
function~\citep{Cox:1979}.

With this Equation in hand, we compute noisy unbiased gradients of the
ELBO with Monte Carlo samples from the variational distribution,
\begin{align}
  \nabla_\lambda \cL \approx \frac{1}{S} \sum_{s=1} ^ S \nabla_\lambda
  \log q(z_s | \lambda) (\log p(x, z_s) - \log q(z_s | \lambda)), \nonumber \\
  \textrm{ where } z_s \sim q(z | \lambda). \label{eq:noisy-gradient}
\end{align}
With \myeqp{noisy-gradient}, we can use stochastic optimization to
optimize the ELBO.

The basic algorithm is summarized in Algorithm \ref{alg:basic}.  We
emphasize that the score function and sampling algorithms depend only
on the variational distribution, not the underlying model. Thus we can
easily build up a collection of these functions for various
variational approximations and reuse them in a package for a variety
of models.  Further we did not make any assumptions about the form of
the model, only that the practitioner can compute the log of the
joint $p(x, z_s)$.  This algorithm significantly reduces the effort
needed to implement variational inference in a wide variety of models.

\begin{algorithm}[tb]
   \caption{Black Box Variational Inference}
   \label{alg:basic}
\begin{algorithmic}
 \STATE {\bfseries Input:} data $x$, joint distribution $p$, mean field variational family $q$.
 \STATE {\bfseries Initialize} $\lambda_{1:n}$ randomly, $t = 1$.
  \REPEAT

  \STATE  {\bfseries // Draw $S$ samples from $q$}
   \FOR{$s=1$ {\bfseries to} S}
     \STATE $z[s]\sim q$
   \ENDFOR
   \STATE $\rho$ = $t$th value of a Robbins Monro sequence (\myeqp{rm})
   \STATE $\lambda$ = $\lambda + \rho \frac{1}{S} \sum_{s=1}^S \nabla_{\lambda} \log q(z[s] | \lambda) (\log p(x, z[s]) - \log q(z[s] | \lambda))$
   \STATE $t = t+1$ 
 \UNTIL{change of $\lambda$ is less than 0.01.}
\end{algorithmic}
\vskip -0.05in
\end{algorithm}

\section{Controlling the Variance}

We can use Algorithm \ref{alg:basic} to maximize the ELBO, but the
variance of the estimator of the gradient (under the Monte Carlo
estimate in \myeqp{noisy-gradient}) can be too large to be useful.  In
practice, the high variance gradients would require very small steps
which would lead to slow convergence.  We now show how to reduce this
variance in two ways, via Rao-Blackwellization and easy-to-implement
control variates.  We exploit the structure of our problem to use
these methods in a way that requires no model-specific derivations,
which preserves our goal of black-box variational inference.

\subsection{Rao-Blackwellization}

Rao-Blackwellization~\citep{Casella:1996} reduces the variance of a
random variable by replacing it with its conditional expectation with
respect to a subset of the variables. Note that the conditional
expectation of a random variable is a random variable with respect to
the conditioning set.  This generally requires analytically computing
problem-specific integrals.
Here we show how to Rao-Blackwellize the estimator for each component
of the gradient without needing to compute model-specific integrals.

In the simplest setting, Rao-Blackwellization replaces a function of
two variables with its conditional expectation. Consider two random
variables, $X$ and $Y$, and a function $J(X, Y)$.  Our goal is to
compute its expectation $\E[J(X,Y)]$ with respect to the joint
distribution of $X$ and $Y$.

Define $\hat{J}(X) = \E[J(X, Y) | X]$, and note that $\E[\hat{J}(X)] =
\E[J(X,Y)]$. This means that $\hat{J}(X)$ can be used in place of
$J(X,Y)$ in a Monte Carlo approximation of $\E[J(X,Y)]$. The variance
of $\hat{J}(X)$ is
\begin{align*}
  \text{Var}(\hat{J}(X)) = \text{Var}(J(X,Y)) - \E[(J(X,Y) -
  \hat{J}(X))^2].
\end{align*}
This means that $\hat{J}(X)$ is a lower variance estimator than
$J(X,Y)$.

We return to the problem of estimating the gradient of ${\cal L}$.
Suppose there are $n$ latent variables $z_{1:n}$ and we are using the
mean-field variational family, where each random variable $z_i$ is
independent and governed by its own variational distribution,
\begin{equation}
  q(z \g \lambda) = \textstyle \prod_{i=1}^{n} q(z_i \g \lambda_i),
\end{equation}
where $\lambda_{1:n}$ are the $n$ variational parameters
characterizing the member of the variational family we seek.  Consider
the $i$th component of the gradient.  Let $q_{(i)}$ be the
distribution of variables in the model that depend on the $i$th
variable, i.e., the Markov blanket of $z_i$; and let $p_i(x,z_{(i)})$
be the terms in the joint that depend on those variables. We can write
the gradient with respect to $\lambda_i$ as an iterated
conditional expectation which simplifies to
\begin{align}
  & \hspace{-5pt} \nabla_{\lambda_i} \cL = \nonumber \\
  & \hspace{5pt} E_{q_{(i)}}[\nabla_{\lambda_i} \log q(z_i | \lambda_i) (\log p_i(x,
  z_{(i)}) - \log q(z_i | \lambda_i))] \label{eq:comp_gradient}.
\end{align}
The derivation of this expression is in the supplement. This equation
says that Rao-Blackwellized estimators can be computed for each
component of the gradient without needing to compute model-specific
conditional expectations.

Finally, we construct a Monte Carlo estimator for the gradient of
$\lambda_i$ using samples from the variational distribution,
\begin{align}
  \frac{1}{S} \sum_{s=1} ^ S \nabla_{\lambda_i} \log q_i(z_s |
  \lambda_i) (\log p_{i}(x, z_s) - \log q_i(z_s | \lambda_i)),
  \nonumber \\
  \textrm{ where } z_s \sim q_{(i)}(z |
  \lambda). \label{eq:factored-estimator}
\end{align}
This Rao-Blackwellized estimator for each component of the gradient
has lower variance. In our empirical study, \myfig{var-red}, we plot the variance of
this estimator along with that of \myeqp{noisy-gradient}.

\subsection{Control Variates}

As we saw above, variance reduction methods work by replacing the
function whose expectation is being approximated by Monte Carlo with
another function that has the same expectation but smaller variance.
That is, to estimate $\E_q[f]$ via Monte Carlo we compute the
empirical average of $\hat{f}$ where $\hat{f}$ is chosen so $\E_q[f] =
E_q[\hat{f}]$ and $\textrm{Var}_q[f] > \textrm{Var}_q[\hat{f}]$.

A control variate~\citep{Ross:2002} is a family of functions with
equivalent expectation.  Consider a function $h$, which has a finite
first moment, and a scalar $a$.  Define $\hat{f}$ to be
\begin{align} \label{eq:arbitrary-cv}
\hat{f}(z) \triangleq f(z) - a (h(z) - E[h(z)]).
\end{align}
This is a family of functions, indexed by $a$, and note that
$E_q[\hat{f}(z)] = \E_q[f]$ as required.  Given a particular function
$h$, we can choose $a$ to minimize the variance of $\hat{f}$.

First we note that variance of $\hat{f}$ can be written as
\begin{align}
\textrm{Var}(\hat{f}) = \textrm{Var}(f) + a^2 \textrm{Var}(h) - 2 a \textrm{Cov}(f, h). \nonumber
\end{align}
This equation implies that good control variates have high covariance
with the function whose expectation is being computed.

Taking the derivative of $\textrm{Var}(\hat{f})$ with respect to $a$
and setting it equal to zero gives us the value of $a$ that minimizes
the variance,
\begin{align}
  a^* = \textrm{Cov}(f, h)/\textrm{Var}(h). \nonumber
\end{align}
With Monte Carlo estimates from the distribution, which we are
collecting anyway to compute $\E[f]$, we can estimate $a^*$ with the
ratio of the empirical covariance and variance.

We now apply this method to Black Box Variational Inference.  To
maintain the generic nature of the algorithm, we want to choose a
control variate that only depends on the variational distribution and
for which we can easily compute its expectation.  Meeting these
criteria, we choose $h$ to be the score function of the variational
approximation, $\nabla_{\lambda} \log q(z)$, which always has
expectation zero.  (See \myeqp{score-zero} in the appendix.)

With this control variate, we have a new Monte Carlo method to compute
the Rao-Blackwellized noisy gradients of the ELBO.  For the $i$th
component of the gradient, the function whose expectation is being
estimated and its control variate, $f_i$ and $h_i$ respectively are
\begin{align}
  \textstyle f_i(z) &= \nabla_{\lambda_i} \log q(z | \lambda_i) (\log p(x, z) - \log q(z | \lambda_i)) \\ \nonumber
  h_i(z) &= \nabla_{\lambda_i} \log q(z | \lambda_i) \nonumber.
\end{align}
The estimate for the optimal choice for the scaling is given by
summing over the covariance and variance for each of the $n_i$
dimensions of $\lambda_i$. Letting the $d$th dimension of $f_i$ and
$h_i$ be $f_i^d$ and $h_i^d$ respectively. The optimal scaling for the
gradient of the ELBO is given by
\begin{align}
  \hat{a}_i^* = \frac{\sum_{d=1}^{n_i} \hat{\textrm{Cov}}(f_i^d,
    h_i^d)} {\sum_{d=1}^{n_i}
    \hat{\textrm{Var}}(h_i^d)}.  \label{eq:opt-cv}
\end{align}

This gives us the following Monte Carlo method to compute noisy gradients using $S$ samples
\begin{align}
\hat{\nabla}_{\lambda_i} \cL \triangleq \frac{1}{S} \sum_{s=1} ^ S &\nabla_\lambda \log q_i(z_s | \lambda_i) 
\nonumber \\
(\log &p_i(x, z_s) - \log q_i(z_s | \lambda_i) - \hat{a}_i^*), 
\nonumber \\
&\text{ where } z_s \sim q_{(i)}(z | \lambda). \label{eq:good-noisy-gradient}
\end{align}
Again note that we define the control variates on a per-component
basis. This estimator uses both Rao-Blackwellization and control
variates.  We show in the empirical study that this generic control
variate further reduces the variance of the estimator.

\subsection{Black Box Variational Inference (II)}

Putting together the noisy gradient, Rao-Blackwellization, and control
variates, we present Black Box Variational Inference (II).  It takes
samples from the variational approximation to compute noisy gradients
as in \myeqp{good-noisy-gradient}. These noisy gradients are then used
in a stochastic optimization procedure to maximize the ELBO.

We summarize the procedure in Algorithm \ref{alg:simple}.  Note that
for simplicity of presentation, this algorithm stores all the samples.
We can remove this memory requirement with a small modification,
computing the $a_i^*$ terms on small set of examples and computing the
required averages online.

Algorithm \ref{alg:simple} is easily used on many models.  It only requires
samples from the variational distribution, computations about the
variational distribution, and easy computations about the model.

\begin{algorithm}[tb]
   \caption{Black Box Variational Inference (II)}
   \label{alg:simple}
\begin{algorithmic}
 \STATE {\bfseries Input:} data $x$, joint distribution $p$, mean field variational family $q$.
 \STATE {\bfseries Initialize} $\lambda_{1:n}$ randomly, $t = 1$.
  \REPEAT

  \STATE  {\bfseries // Draw $S$ samples from the variational approximation}
   \FOR{$s=1$ {\bfseries to} S}
     \STATE $z[s]\sim q$
   \ENDFOR
   \FOR{$i=1$ {\bfseries to} n}
   	\FOR{$s=1$ {\bfseries to} S}
		\STATE $f_i[s] = \nabla_{\lambda_i} \log q_i(z[s] | \lambda_i) (\log p_i(x, z[s]) - \log q_i(z[s] | \lambda_i))$
		\STATE $h_i[s] = \nabla_{\lambda_i} \log q_i(z[s] | \lambda_i)$
	\ENDFOR
     \STATE $\hat{a_i^*} = \frac{\sum_{d=1}^{n_i} \hat{\textrm{Cov}}(f_i^d, h_i^d)} 
		{\sum_{d=1}^{n_i} \hat{\textrm{Var}}(h_i^d)}$
     \STATE $\hat{\nabla}_{\lambda_i} \cL \triangleq \frac{1}{S} \sum_{s=1} ^ S f_i[s] - \hat{a_i^*} h_i[s]$
   \ENDFOR
   \STATE $\rho$ = $t$th value of a Robbins Monro sequence
   \STATE $\lambda = \lambda + \rho \hat{\nabla}_{\lambda} \cL$
   \STATE $t = t+1$ 
 \UNTIL{change of $\lambda$ is less than 0.01.}
\end{algorithmic}
\vskip -0.05in
\end{algorithm}

\section{Extensions}

We extend the main algorithm in two ways. First, we address the
difficulty of setting the step size schedule. Second, we address
scalability by subsampling observations.

\subsection{AdaGrad}

One challenge with stochastic optimization techniques is setting the
learning rate. Intuitively, we would like the learning rate to be
small when the variance of the gradient is large and vice-versa.
Additionally, in problems like ours that have different
scales\footnote{Probability distributions have many
  parameterizations.}, the learning rate needs to be set small enough
to handle the smallest scale. To address this issue, we use the
AdaGrad algorithm ~\citep{Duchi:2011}.  Let $G_t$ be a matrix
containing the sum across the first $t$ iterations of the outer
products of the gradient.  AdaGrad defines a per component learning
rate as
\begin{align}
\rho_t = \eta \textrm{diag}(G_t)^{-1/2}.
\end{align}
This is a per-component learning rate since $\textrm{diag}(G_t)$ has
the same dimension as the gradient. Note that since AdaGrad only uses
the diagonal of $G_t$, those are the only elements we need to compute.
AdaGrad captures noise and varying length scales through the square of
the noisy gradient and reduces the number of parameters to our
algorithm from the standard two parameter Robbins-Monro learning rate.

\subsection{Stochastic Inference in Hierarchical Bayesian Models}
Stochastic optimization has also been used to scale variational
inference in hierarchical Bayesian models to massive
data~\citep{Hoffman:2013}.  The basic idea is to subsample
observations to compute noisy gradients.  We can use a similar idea to
scale our method.

In a hierarchical Bayesian model, we have a
hyperparameter $\eta$, global latent variables $\beta$, local latent
variables $z_{1...n}$, and observations $x_{1...n}$ having the log joint
distribution
\begin{align} \label{eq:h-bayes-model}
\log p (x_{1...n}, z_{1...n}, \beta) =& \log p(\beta | \eta) \nonumber \\
&+ \sum_{i = 1}^n \log p(z_i | \beta) + \log p (x_i | z_i, \beta).
\end{align}
This is the same definition as in \citet{Hoffman:2013}, but they place
further restrictions on the forms of the distributions and the complete
conditionals.  Under the mean field approximating family, applying
\myeqp{good-noisy-gradient} to construct noisy gradients of the ELBO
would require iterating over every datapoint. Instead we can compute
noisy gradients using a sampled observation and samples from the
variational distribution. The derivation along with variance
reductions can be found in the supplement.

\section{Empirical Study} \label{sec:experiments}

We use Black Box Variational Inference to quickly construct and
evaluate several models on longitudinal medical data. We demonstrate
the effectiveness of our variance reduction methods and compare the
speed and predictive likelihood of our algorithm to sampling based
methods.  We evaluate the various models using predictive likelihood
and demonstrate the ease at which several models can be explored.

\subsection{Longitudinal Medical Data}

Our data consist of longitudinal data from 976 patients (803 train +
173 test) from a clinic at New York Presbyterian hospital who have
been diagnosed with chronic kidney disease. These patients visited the
clinic a total of 33K times.  During each visit, a subset of 17
measurements (labs) were measured.

The data are observational and consist of measurements (lab values)
taken at the doctor's discretion when the patient is at a
checkup. This means both that the labs at each time step are sparse
and that the time between patient visits are highly irregular. The
labs values are all positive as the labs measure the amount of a
particular quantity such as sodium concentration in the blood.

Our modeling goal is to come up with a low dimensional summarization
of patients' labs at each of their visits. From this, we aim to to
find latent factors that summarize each visit as positive random
variables.  As in medical data applications, we want our factors to be
latent indicators of patient health.

We evaluate our model using predictive likelihood.  To compute
predictive likelihoods, we need an approximate posterior on both the
global parameters and the per visit parameters. We use the approximate
posterior on the global parameters and calculate the approximate
posterior on the local parameters on 75\% of the data in the test
set. We then calculate the predictive likelihood on the other 25\% of
the data in the validation set using Monte Carlo samples from the
approximate posterior.

We initialize randomly and choose the variational families to be
fully-factorized with gamma distributions for positive variables and normals 
for real valued variables.
We use both the AdaGrad and
doubly stochastic extensions on top of our base algorithm. We use
1,000 samples from the variational distribution and set the batch size
at 25 in all our experiments.

\subsection{Model}

\begin{figure}[t]
  \begin{center}
\includegraphics[width=.5\textwidth]{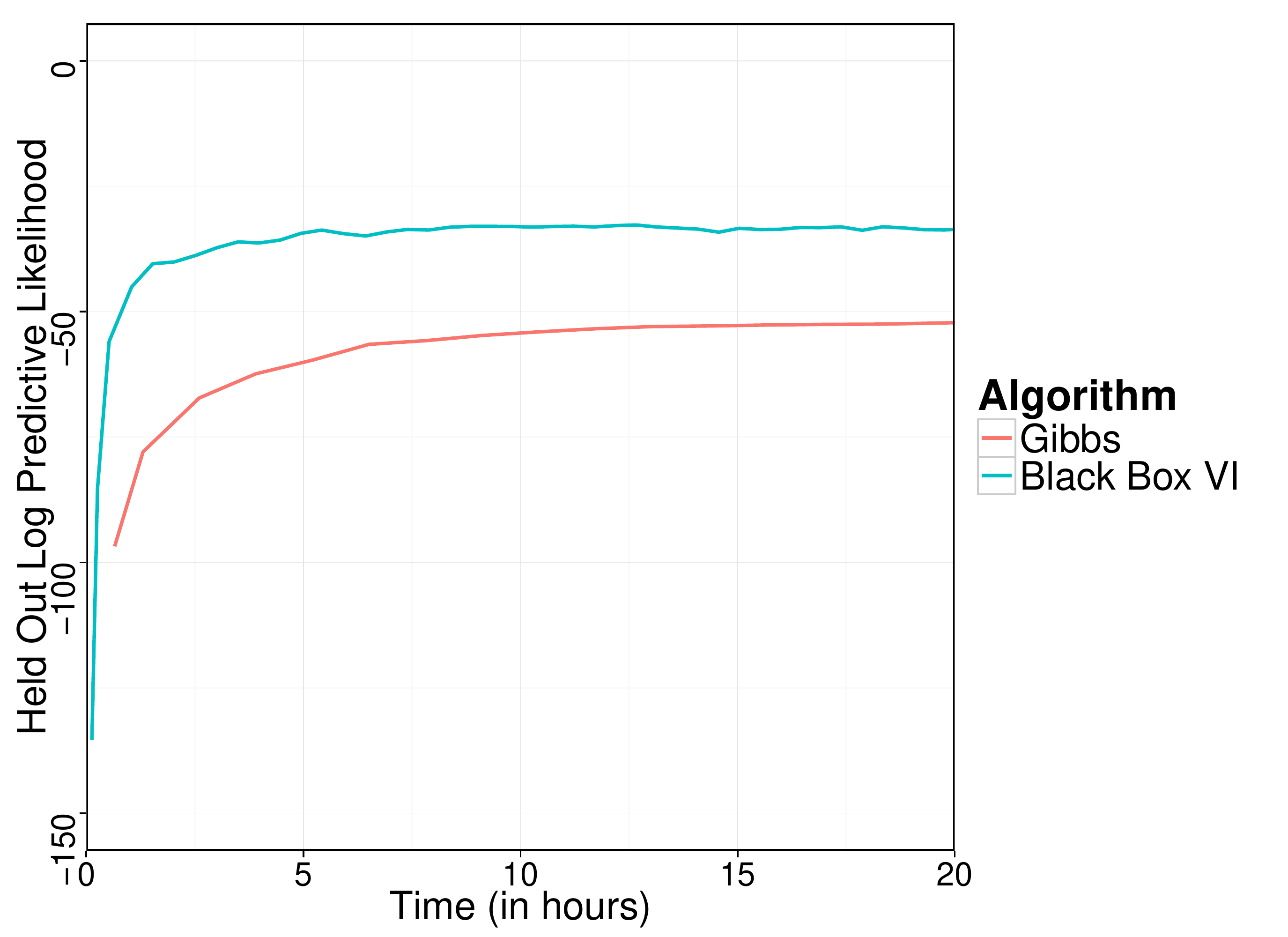}
\vskip -0.05in
\caption{Comparison between Metropolis-Hastings within Gibbs and Black
  Box Variational Inference.  In the x axis is time and in the y axis
  is the predictive likelihood of the test set.  Black Box Variational
  Inference reaches better predictive likelihoods faster than Gibbs
  sampling. The Gibbs sampler's progress slows considerably after 5
  hours.}
\label{fig:gibbs-vi}
  \end{center}
\end{figure}

To meet our goals, we construct a Gamma-Normal time series model.  We
model our data using weights drawn from a Normal distribution and
observations drawn from a Normal, allowing each factor to both
positively and negative affect each lab while letting factors
represent lab measurements. The generative process for this model with
hyperparameters denoted with $\sigma$ is
\begin{tabbing}
D\=raw $W$ $\sim$ Normal$(0, \sigma_w)$, an $L \times K$ matrix \\
For each patient $p$: $1$ to $P$ \\
\> Draw $o_p$ $\sim $Normal$(0, \sigma_o)$, a vector of $L$\\
\> Define $x_{p0} = \alpha_0$ \\
\> F\=or each visit $v$: 1 to $v_p$ \\
\> \> Draw $x_{pv}$ $\sim $GammaE$(x_{pv-1}, \sigma_x)$ \\
\> \> Draw $l_{pv}$ $\sim $Normal$(Wx_{pv} + o_p, \sigma_l)$, a vector of $L$.
\end{tabbing}

We set $\sigma_w$, $\sigma_o$, $\sigma_x$ to be 1 and $\sigma_l$ to be
.01. In our model, GammaE is the expectation/variance parameterization
of the (L-dimensional) gamma distribution. (The mapping between this
parameterization and the more standard one can be found in the
supplement.)

Black Box Variational Inference allows us to make use of non-standard
parameterizations for distributions that are easier to reason about.
This is an important observation, as the standard set of families used
in variational inference tend to be fairly limited. In this case, the
expectation parameterization of the gamma distribution allows the
previous visit factors to define the expectation of the current visit
factors.  Finally, we emphasize that coordinate ascent variational
inference and Gibbs sampling are not available for this algorithm
because the required conditional distributions do not have closed
form.

\subsection{Sampling Methods}
We compare Black Box Variational Inference to a standard sampling
based technique, Metropolis-Hastings~\citep{Bishop:2006}, that also
only needs the joint distribution.\footnote{Methods that involve a bit
  more work such as Hamiltonian Monte Carlo could work in this
  settings, but as our technique only requires the joint distribution
  and could benefit from added analysis used in more complex methods,
  we compare against a similar methods.}
 
Metropolis-Hastings works by sampling from a proposal distribution and
accepting or rejecting the samples based on the likelihood.  Standard
Metropolis-Hastings can work poorly in high dimensional models. We
find that it fails for the Gamma-Normal-TS model.  Instead, we compare
to a Gibbs sampling method that uses Metropolis-Hastings to sample
from the complete conditionals.  For our proposal distribution we use
the same distributions as found in the previous iteration, with the
mean equal the value of the previous parameter.

We compute predictive likelihoods using the posterior samples generated
by the MCMC methods on held out data in the test set.

On this model, we compared Black Box Variational Inference to
Metropolis-Hastings inside Gibbs.  We used a fixed computational
budget of 20 hours. \myfig{gibbs-vi} plots time versus predictive
likelihood on the held out set for both methods. We found similar
results with different random initializations of both models. Black
Box Variational Inference gives better predictive likelihoods and gets
them faster than Metropolis-Hastings within Gibbs.\footnote{Black Box
  Variational Inference also has better predictive mean-squared error on
the labs than Gibbs style Metropolis-Hastings.}.

\subsection{Variance Reductions}
\begin{figure}[t]
  \begin{center}
\includegraphics[width=.5\textwidth]{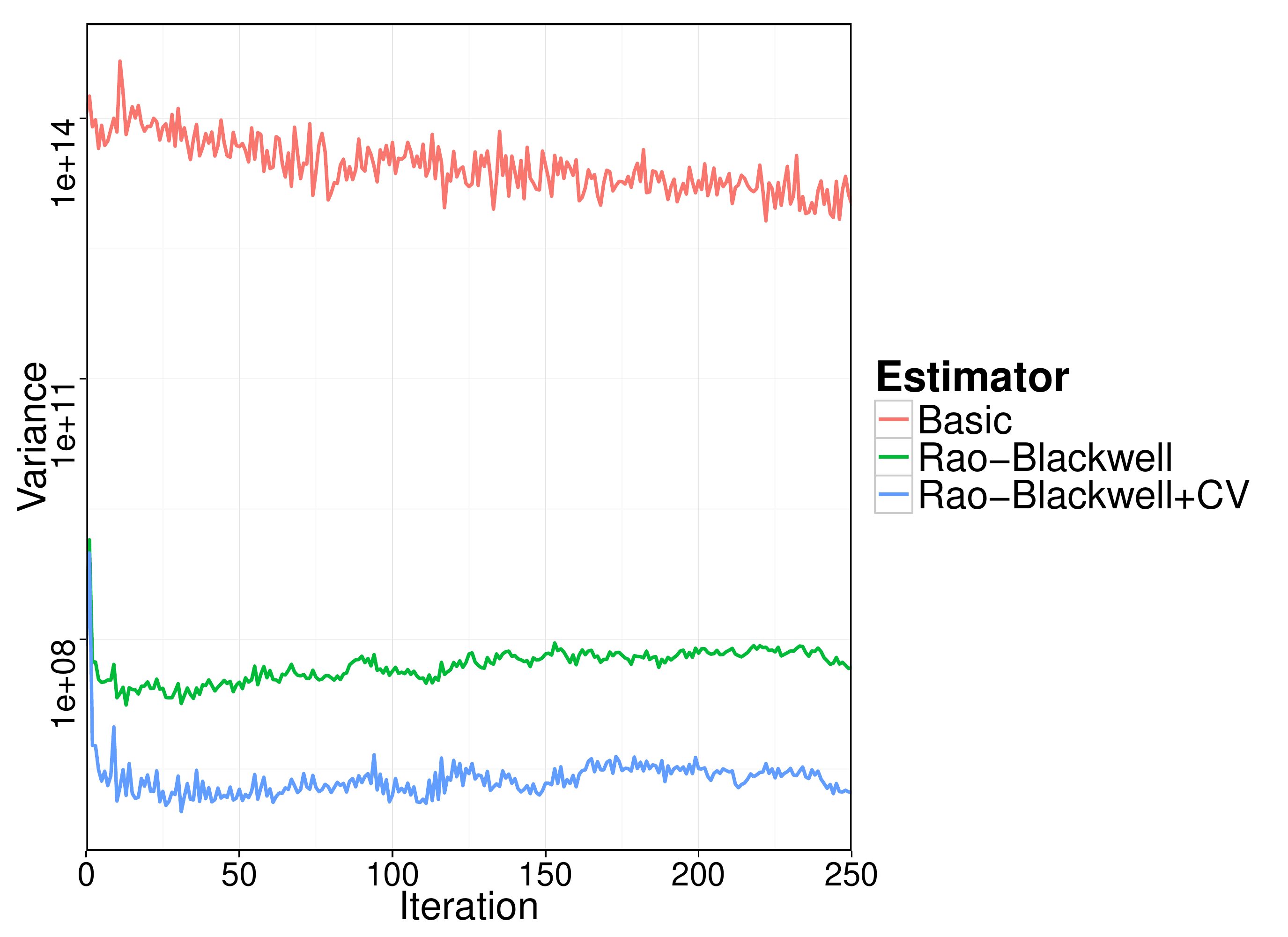}
\vskip -0.05in
\caption{Variance comparison for the first component of a random patient
  on the following estimators:
  \myeqp{noisy-gradient}, the Rao-Blackwellized estimator
  \myeqp{factored-estimator}, and the Rao-Blackwellized control
  variate estimator \myeqp{good-noisy-gradient}. We find that
  Rao-Blackwellizing the naive estimator reduces the variance by
  several orders of magnitude from the naive estimator. 
  Adding control variates reduces the variance even further. }
\label{fig:var-red}
  \end{center}
\end{figure}

We next studied how much variance is reduced with our variance
reduction methods.  In \myfig{var-red}, we plot the variance of
various estimators of the gradient of the variational approximation
for a factor in the patient time-series versus iteration number.  We
compare the variance of the Monte Carlo gradient in
~\myeqp{noisy-gradient} to that of the Rao-Blackwellized
gradient (\myeqp{factored-estimator}) and that of the gradient using
both Rao-Blackwellization and control variates
(\myeqp{good-noisy-gradient}).  We found that Rao-Blackwellization
reduces the variance by several orders of magnitude. Applying control
variates reduces the variance further. This reduction in variance
drastically improves the speed at which Black Box Variational
Inference converges. In fact, in the time allotted,
Algorithm~\ref{alg:basic}---the algorithm without variance
reductions---failed to make noticeable progress.

\subsection{Exploring Models}

We developed Black Box Variational Inference to make it easier to
quickly explore and fit many new models to a data set.  We demonstrate
this by considering a sequence of three other factor and time-series
models of the health data.  We name these models Gamma, Gamma-TS, and
Gamma-Normal.

\paragraph{Gamma.}  We model the latent factors that summarize each
visit in our models as positive random variables; as noted above, we
expect these to be indicative of patient health.  The Gamma model is a
positive-value factor model where all of the factors, weights, and
observations have positive values. The generative process for this
model is
\begin{tabbing}
D\=raw $W$ $\sim$ Gamma$(\alpha_w, \beta_w)$, an $L \times K$ matrix \\
For each patient $p$: $1$ to $P$ \\
\> Draw $o_p$ $\sim $Gamma$(\alpha_o, \beta_o)$, a vector of $L$ \\
\> F\=or each visit $v$: 1 to $v_p$ \\
\> \> Draw $x_{pv}$ $\sim $Gamma$(\alpha_x, \beta_x)$ \\
\> \> Draw $l_{pv}$ $\sim $GammaE$(Wx_{pv} + o_p, \sigma_o)$, a vector of $L$.
\end{tabbing}
We set all hyperparameters save $\sigma_o$ to be 1. As in the previous
model, $\sigma_o$ is set to .01.

\paragraph{Gamma-TS.}
We can link the factors through time using the expectation
parameterization of the Gamma distribution.  (Note this is harder with
the usual natural parameterization of the Gamma.)  This changes
$x_{pv}$ to be distributed as GammaE$(x_{pv-1}, \sigma_v)$.  We draw
$x_{p1}$ as above. In this model, the expected values of the factors
at the next visit is the same as the value at the current visit. This
allows us to propagate patient states through time. 

\paragraph{Gamma-Normal.}
Similar to the above, we can change the time-series Gamma-Normal-TS
(studied in the previous section) to a simpler factor model.  This is
similar to the Gamma model, but with Normal priors.

These combinations lead to a set of four models that are all
nonconjugate and for which standard variational techniques are
difficult to apply. Our variational inference method allows us to
compute approximate posteriors for these models to determine which
provides the best low dimensional latent representations.

We set the AdaGrad scaling parameter to $1$ for both the Gamma-Normal models
and to $.5$ for the Gamma models.

\paragraph{Model Comparisons.}
Table \ref{tab:results_model_comp} details our models along with their
predictive likelihoods.  From this we see that time helps in modelling
our longitudinal healthcare data. We also see that the Gamma-Gamma
models perform poorly. This is likely because they cannot capture the
negative correlations that exist between different medical labs.  More
importantly, by using Black Box Variational Inference we were able to
quickly fit and explore a set of complicated non-conjugate models.
Without a generic algorithm, approximating the posterior of any of
these models is a project in itself.

\begin{table}
\centering
\begin{tabular}{|c | c |}
\hline 
 Model & Predictive Likelihood  \\ \hline
 Gamma-Normal & -33.9 \\
 Gamma-Normal-TS  & -32.7 \\
 Gamma-Gamma & -175 \\
 Gamma-Gamma-TS  & -174 \\
\hline  
\end{tabular}
\caption{A comparison between several models for
our patient health dataset. We find that taking into 
account the longitudinal nature of the data in the
model leads to a better fit. The Gamma weight models
perform relatively poorly. This is likely due to the fact
that some labs have are negatively correlated. This
model cannot capture such relationships.
}  \label{tab:results_model_comp}
\end{table}

\section{Conclusion}

We developed and studied Black Box Variational Inference, a new
algorithm for variational inference that drastically reduces the
analytic burden.  Our main approach is a stochastic optimization of
the ELBO by sampling from the variational posterior to compute a noisy
gradient.  Essential to its success are model-free variance reductions
to reduce the variance of the noisy gradient. Black Box Variational
Inference works well for new models, while requiring minimal analytic
work by the practitioner.

There are several natural directions for future improvements to this
work.  First, the software libraries that we provide can be augmented
with score functions for a wider variety of variational families (each
score function is simply the log gradient of the variational
distribution with respect to the variational parameters). Second, we
believe that number of samples could be set dynamically.  Finally,
carefully-selected samples from the variational distribution (e.g.,
with quasi-Monte Carlo methods) are likely to significantly decrease
sampling variance.

\section{Appendix: The Gradient of the ELBO}
The key idea behind our algorithm is that the gradient of the ELBO can
be written as an expectation with respect to the variational
distribution.  We start by differentiating \myeqp{objective},
\begin{eqnarray}
  \nabla_\lambda \cL &=& \nabla_\lambda \int (\log p(x, z) - \log q(z |
  \lambda)) q(z | \lambda) dz \nonumber \\
    \nonumber &=& \int \nabla_\lambda [(\log p(x, z) - \log q(z |
  \lambda)) q(z | \lambda)] dz \\
  \label{eq:ugly-gradient} &=& \int \nabla_\lambda [\log p(x, z) - \log
  q(z | \lambda)] q(z | \lambda)dz \nonumber \\
   &&+ \int \nabla_\lambda q(z |
  \lambda) (\log p(x, z) - \log q(z | \lambda)) dz \nonumber \\
  &=& \label{eq:ugly-gradient}
  - \E_q[\log q(z | \lambda)]  \\ \nonumber
&&+ \int \nabla_\lambda q(z |
  \lambda) (\log p(x, z) - \log q(z | \lambda)) dz,
\end{eqnarray}
where we have exchanged derivatives with integrals via the dominated
convergence theorem \footnote{The score function exists. The score and
  likelihoods are bounded.} \citep{Cinlar:2011} and used
$\nabla_\lambda [\log p(x, z)] = 0$.

The first term in \myeqp{ugly-gradient} is zero.  To see this, note
\begin{align}
\E_q[\nabla_\lambda \log q(z | \lambda)] &=
\E_q\left[\frac{\nabla_\lambda q(z | \lambda)}{q(z | \lambda)}\right]
= \int \nabla_\lambda q(z | \lambda) dz \nonumber \\
&= \nabla_\lambda \int q(z |
\lambda) dz = \nabla_\lambda 1 = 0. \label{eq:score-zero}
\end{align}

To simplify the second term, first observe that $\nabla_\lambda [q(z |
\lambda)] = \nabla_\lambda [\log q(z | \lambda)] q(z | \lambda)$.
This fact gives us the gradient as an expectation,
\begin{eqnarray}
  \nabla_\lambda \cL &=& \int \nabla_\lambda [q(z | \lambda)] (\log
  p(x, z) - \log q(z | \lambda)) dz \nonumber \\
  &=& \int \nabla_\lambda \log q(z | \lambda) (\log p(x, z) \nonumber \\
&&- \log q(z
  | \lambda)) q(z | \lambda) dz   \nonumber
 \\ \nonumber
  &=& \E_q [\nabla_\lambda \log q(z | \lambda) (\log p(x, z) - \log q(z | \lambda))],
\end{eqnarray}

\bibliography{arxiv}

\section*{Supplement}
\renewcommand{\theequation}{A.\arabic{equation}}
\paragraph{Derivation of the Rao-Blackwellized Gradient} 
To compute the Rao-Blackwellized estimators, we 
need to compute conditional expectations.
Due to the mean
field-assumption, the conditional expectation simplifies due to the
factorization
\begin{align}
 \E[J(X, Y) | X] &= \frac{\int J(x, y) p(x) p(x) dy}{\int p(x) p(y) dy} \nonumber \\
  &= \int J(x, y) p(y) dy = \E_y[J(x, y)]. \label{eq:rao-blackwell}
\end{align}
 Therefore, to construct a lower variance estimator when the joint
 distribution factorizes, all we need to do is integrate out some
 variables.  In our problem this means for each component of the
 gradient, we should compute expectations with respect to the other
 factors.  We present the estimator in the full mean field family of
 variational distributions, but note it applies to any variational
 approximation with some factorization like structured mean-field.

Thus, under the 
mean field assumption the Rao-Blackwellized
estimator for the gradient
becomes
\begin{align}
\nabla_\lambda \cL =& \E_{q_1} \ldots \E_{q_n} [\sum_{j=1}^n \nabla_\lambda \log q_j(z_j | \lambda_j) (\log p(x, z)
\nonumber \\
 &- \sum_{j=1}^n \log q_j(z_j | \lambda_j))].
\end{align}
Recall the definitions from Section 3 where we defined 
$\nabla_{\lambda_i} \cL$ as the gradient of the ELBO with
respect to $\lambda_i$, $p_{i}$ as the 
components of the log joint that include terms form the
$i$th factor, and $E_{q_{(i)}}$ as
the expectation with respect to the set of latent variables
that appear in the complete conditional for $z_i$. Let
$p_{-i}$ bet the
components of the joint that does not include terms 
from the $i$th factor respectively. We can write the 
gradient with respect to the $i$th factor's variational parameters as
\begin{align}
\nabla_{\lambda_i}& \cL 
\nonumber \\
=& E_{q_1} \ldots \E_{q_n} [\nabla_{\lambda_i} \log q_i(z_i | \lambda_i) (\log p(x, z) 
\nonumber \\
&- \sum_{j=1}^n \log q_j(z_j | \lambda_j))] \nonumber
\\ \nonumber
=& E_{q_1} \ldots \E_{q_n} [\nabla_{\lambda_i} \log q_i(z_i | \lambda_i) (\log p_{i}(x, z) 
\\ \nonumber
&+ \log p_{-i}(x, z) - \sum_{j=1}^n \log q_j(z_j | \lambda_j))]
\\ \nonumber
=& E_{q_i} [\nabla_{\lambda_i} \log q_i(z_i | \lambda_i) (E_{q_{-i}} [\log p_{i}(x, z_{(i)})] 
\\ \nonumber
&- \log q_i(z_i | \lambda_i) + \E_{q_{-i}} [\log p_{-i}(x, z)
\\ \nonumber
&- \sum_{j=1, i \neq j}^n \log q_j(z_j | \lambda_j)]]
\\ \nonumber
=& E_{q_i} [\nabla_{\lambda_i} \log q_i(z_i | \lambda_i) (\E_{q_{-i}} [\log p_{i}(x, z)] 
\\ \nonumber
&- \log q_i(z_i | \lambda_i) + C_i)]
\\ \nonumber
=& E_{q_i} [\nabla_{\lambda_i} \log q_i(z_i | \lambda_i) (\E_{q_{-i}}[\log p_{i}(x, z_{(i)})] 
\\ \nonumber
&- \log q_i(z_i | \lambda_i))]
\\ \label{eq:comp_gradient}
=& E_{q_{(i)}}[\nabla_{\lambda_i} \log q_i(z_i | \lambda_i) (\log p_{i}(x, z_{(i)}) - \log q_i(z_i | \lambda_i))].
\end{align}
where we have leveraged the mean field assumption and made use of the identity for 
the expected score Eq. 14. 
This means we can Rao-Blackwellize the gradient of
the variational parameter $\lambda_i$ with respect to the
the latent variables outside of the Markov blanket of $z_i$ without
needing model specific computations.

\paragraph{Derivation of Stochastic Inference in Hierarchical Bayesian Models} Recall 
the definition of a hierarchical Bayesian model with $n$ observations
given in Eq. 12
\begin{align}
\log &p (x_{1...n}, z_{1...n}, \beta) 
\nonumber \\
=& \log p(\beta | \eta) +  \sum_{i = 1}^n \log p(z_i | \beta) + \log p (x_i, | z_i, \beta).
\nonumber
\end{align}
Let the variational approximation for the posterior distribution be
from the mean field family. Let $\lambda$ be the global variational 
parameter and let $\phi_{1...n}$ be the local variational parameters.
The variational family is
\begin{align}
q(\beta, z_{1...n}) = q(\beta | \lambda) \prod_{i=1}^m q(z_i | \phi_i).
\end{align}

Using the Rao Blackwellized estimator to compute noisy gradients
in this family for this model gives
\begin{align}
\hat{\nabla}_{\lambda} \cL =& \frac{1}{S} \sum_{i=1}^S \nabla_\lambda \log q(\beta_s | \lambda) (\log p (\beta_s | \eta) -  \log q(\beta_s | \lambda) 
\nonumber \\
&+ \sum_{i=1}^n (\log p({z_i}_s | \beta_s) + \log p (x_i, {z_i}_s | \beta_s)))
 \nonumber \\
\hat{\nabla}_{\phi_i} \cL =& \frac{1}{S} \sum_{i=1}^S \nabla_\lambda \log q(z_{s} | \phi_i) ((\log p({z_i}_s | \beta_s)
\nonumber \\ &+ \log p (x_i, {z_i}_s | \beta_s)  - \log q ({z_i}_s | \phi_i))).
\nonumber
\end{align}
Unfortunately, this estimator requires iterating over every data point to compute noisy realizations
of the gradient. We can mitigate this by subsampling observations. If we let 
$i \sim Unif(1...n)$, then we can write down a noisy gradient for the ELBO that does
not need to iterate over every observation; this noisy gradient is
\begin{align}
\hat{\nabla}_{\lambda} \cL =& \frac{1}{S} \sum_{i=1}^S \nabla_\lambda \log q(\beta_s | \lambda) (\log p (\beta_s | \eta) - \log q(\beta_s | \lambda) 
\nonumber \\
&- n (\log p({z_i}_s | \beta_s) + \log p (x_i, {z_i}_s | \beta_s)))
 \nonumber \\
\hat{\nabla}_{\phi_i} \cL =& \frac{1}{S} \sum_{i=1}^S \nabla_\lambda \log q(z_{s} | \phi_i) (n (\log p({z_i}_s | \beta_s) 
\nonumber \\
&+ \log p (x_i, {z_i}_s | \beta_s) - \log q ({z_i}_s | \phi_i)))
\nonumber \\
\hat{\nabla}_{\phi_j} \cL =& 0 \textrm{ for all $j \neq i$}. \nonumber
\end{align}
The expected value of this estimator with respect to the samples from the variational distribution
and the sampled data point is the gradient of the ELBO. This means we can use it define a stochastic
optimization procedure to maximize the ELBO.  We can lower the variance of the estimator by 
introducing control variates. Let
\begin{align}
  \textstyle f_\lambda(\beta, z_i) =& \nabla_{\lambda} \log q(\beta | \lambda) (\log p (\beta_s | \eta) - \log q(\beta_s | \lambda) 
\nonumber \\
&- n (\log p({z_i}_s | \beta_s) + \log p (x_i, {z_i}_s | \beta_s)))  \nonumber \\
  h_\lambda(\beta) =& \nabla_{\lambda} \log q(\beta | \lambda_i)  \nonumber \\ 
  \textstyle f_{\phi_i}(\beta, z_i) =& \nabla_{\lambda_i} \log q(z | \lambda_i) (n (\log p({z_i}_s | \beta_s) 
\nonumber \\
&+ \log p (x_i, {z_i}_s | \beta_s) - \log q ({z_i}_s | \phi_i))) \nonumber \\ 
  h_{\phi_i}(z_i) =& \nabla_{\lambda_i} \log q(z_i | \phi_i). \label{eq:hbayes-cv}
\end{align}
We can compute the optimal scalings for the control variates, $\hat{a_\lambda^*}$ and $\hat{a_{\phi_i}^*}$, using the $S$ Monte Carlo by substituting Eq.~\ref{eq:hbayes-cv} into Eq. 9.
This gives the following lower variance noisy gradient
that does not need to iterate over all of the observations
at each update
\begin{align}
\hat{\nabla}_{\lambda} \cL =& \frac{1}{S} \sum_{i=1}^S \nabla_\lambda \log q(\beta_s | \lambda) (\log p (\beta_s | \eta) - \log q(\beta_s | \lambda) 
\nonumber \\
&-\hat{a_\lambda^*} + n (\log p({z_i}_s | \beta_s) + \log p (x_i, {z_i}_s | \beta_s)))
 \nonumber \\
\hat{\nabla}_{\phi_i} \cL =& \frac{1}{S} \sum_{i=1}^S \nabla_\lambda \log q(z_{s} | \phi_i) (-\hat{a_{\phi_i}^*} + n (\log p({z_i}_s | \beta_s) 
\nonumber \\
&+ \log p (x_i, {z_i}_s | \beta_s) - \log q ({z_i}_s | \phi_i)))
\nonumber \\
\hat{\nabla}_{\phi_j} \cL =& 0 \textrm{ for all $j \neq i$}.
\end{align}

\paragraph{Gamma parameterization equivalence} The 
shape $\alpha$ and rate $\beta$ parameterization can be written 
in terms of the mean $\mu$ and variance $\sigma^2$ of the gamma
as
\begin{align}
\alpha = \frac{\mu^2}{\sigma^2}, \quad \beta = \frac{\mu}{\sigma^2}.
\end{align}

\end{document}